\documentclass[12pt, final, nosubfloats]{l4dc2023}

\title[]{Continuous Versatile Jumping Using Learned Action Residuals}
\usepackage{times}
\usepackage{amsmath}
\usepackage{amsfonts}
\usepackage{amssymb}
\usepackage{gensymb}
\usepackage{bbm}
\usepackage{bm}
\usepackage[font=small]{caption}
\usepackage{graphicx}
\usepackage{subcaption}
\usepackage{natbib}

\renewcommand{\vec}{\bm}

\author{%
 \Name{Yuxiang Yang} \Email{yuxiangy@cs.washington.edu}\AND
 \Name{Xiangyun Meng} \Email{xiangyun@cs.washington.edu}\AND
 \Name{Wenhao Yu} \Email{magicmelon@google.com}\AND
 \Name{Tingnan Zhang} \Email{tingnan@google.com}\AND
 \Name{Jie Tan} \Email{jietan@google.com}\AND
 \Name{Byron Boots} \Email{bboots@cs.washington.edu}\AND
 \addr  University of Washington\\
 \addr  Robotics at Google%
}

\begin{document}

\maketitle

\begin{abstract}%
Jumping is essential for legged robots to traverse through difficult terrains. 
In this work, we propose a hierarchical framework that combines optimal control and reinforcement learning to learn continuous jumping motions for quadrupedal robots.
The core of our framework is a stance controller, which combines a manually designed acceleration controller with a learned residual policy. As the acceleration controller warm starts policy for efficient training, the trained policy overcomes the limitation of the acceleration controller and improves the jumping stability. 
In addition, a low-level whole-body controller converts the body pose command from the stance controller to motor commands.
After training in simulation, our framework can be deployed directly to the real robot, and perform versatile, continuous jumping motions, including omni-directional jumps at up to 50cm high, 60cm forward, and jump-turning at up to 90 degrees.
Please visit our website for more results: \url{https://sites.google.com/view/learning-to-jump}.
\end{abstract}


\section{Introduction}
Jumping can greatly extend the capabilities of legged robots.
Compared to walking, jumping exhibits a long "air phase", where all legs leave the ground at the same time.
This air phase enables the robot to traverse through large areas without making contact, which is essential for difficult terrains with large gaps or abrupt height changes.
While recent works have greatly improved the speed \citep{mit_cmpc, mit_wbic, margolis2022rapid} and robustness \citep{rma, rma_vision, eth_adaptation} of legged robots, most of them focused on standard walking behaviors with continuous foot contacts.
Meanwhile, long-distance jumping is still a difficult task, and usually requires manual trajectory design \citep{li2022versatile, park2021jumping} as well as long periods of offline planning \citep{mitomnidirectional, towr}.

In this work, we present a hierarchical learning framework for quadrupedal robots to jump continuously, where the jumping direction and distance can be specified online. 
Continuous jumping has long been a difficult task for legged robots due to complex robot dynamics and frequent, abrupt contact changes.
On the one hand, while optimal control based controllers have achieved robust walking in many quadruped platforms \citep{mit_cmpc, grandia2019feedback}, they usually assume a simplified dynamics model for computation efficiency \citep{mit_cmpc}, and cannot control the robot pose precisely in highly dynamic motions like jumping \citep{mitomnidirectional}.
On the other hand, despite recent success in learning for locomotion \citep{rma,margolis2022rapid,eth_adaptation}, reinforcement learning (RL) based controllers still require careful reward tuning \citep{laikago_imitation} and extended training times to learn jumping motions, due to the non-smooth reward landscape created by abrupt contact changes.
Therefore, it can be difficult to use standard control or learning techniques for the jumping task.

Our framework addresses the challenges above, and learns continuous, versatile jumping motions that can be transferred directly to the real world.
The core of our framework is a \emph{stance controller}, which computes desired body pose by summing over the outputs from a manually-designed \emph{acceleration controller} and a learned \emph{residual policy}.
Our design of the stance controller has two major benefits.
Firstly, warm-starting the policy training with the acceleration controller reduces noises in the reward landscape, so that training process converges smoothly and efficiently.
Secondly, with the residual policy trained, the robot performance is no longer limited by the simplified dynamics model used by the acceleration controller, and therefore can better stabilize the robot throughout the entire jumping episode.
In addition to the stance controller, we also implemented a low-level \emph{whole-body controller} to convert the body pose command to motor actions.
By combining the acceleration controller, the residual policy and the low-level whole-body controller, our framework learns continuous, versatile jumping motions automatically.

We train our framework on a simulated environment of the Go1 quadruped robot from Unitree \citep{go1robot}, and test the trained policy directly on the real robot.
The trained framework enables versatile jumping motions for the robot, including jumping at different directions and distances (up to 50cm high, 60cm forward), and jump-turning (up to 90 degrees).
We then conduct detailed analysis on the behavior of our overall framework, and verify that the combination the acceleration controller and residual policy can learn more stable jumping motions than each individual method.
Additionally, we compare our method to end-to-end RL and find that our method is at least 1 order-of-magnitude more data efficient, thanks to the hierarchical setup and the smooth reward landscape from the acceleration controller.

In summary, the contributions of this paper include the following:
\begin{enumerate}
    \item We propose a hierarchical framework that combines optimal control and reinforcement learning to learn continuous, versatile jumping for quadrupedal robots.
    \item The trained framework can be directly transferred to the real robot and achieves continuous jumping motions at substantial height (50cm) and distance (60cm).
    \item Our experiments show that the combination of controller and residual policy can learn more stable jumping motions than using either method individually.
\end{enumerate}

\section{Related Works}
\paragraph{Learning Agile Locomotion}
Recently, researchers have made significant progress in applying reinforcement learning (RL) for quadrupedal locomotion.
Using reinforcement learning, the legged robots can adapt to the environment \citep{rma, eth_adaptation} and learn diverse skills such as self-righting \citep{actuatornet, wu2022daydreamer}, high-speed walking \citep{margolis2022rapid}, goal-keeping \citep{huang2022creating}.
However, for successful real-robot deployment, RL-based controllers usually require significant effort in imitation learning\citep{laikago_imitation}, reward shaping \citep{rma} and sim-to-real \citep{sim-to-real, actuatornet, song2020rapidly}, especially for more agile tasks such as jumping.
Compared to the end-to-end RL approaches, we re-design the RL task to learn the residual action \citep{johannink2019residual} that adds to a manually-designed acceleration controller. As a result, the training process consumes significantly fewer data and does not require complex reward specification. More over, the learned policy can be deployed directly to the real world without additional fine-tuning.

\paragraph{Optimal Control for Locomotion}
With recent advancements in actuator design and numerical optimization, optimal-control based controllers have enabled high-speed and robust locomotion \citep{mit_cmpc, mit_wbic, grandia2019feedback} for legged robots.
For computation efficiency, these controllers usually simplify the robot dynamics model, such as the single rigid body model with massless legs \citep{mit_cmpc, li2022versatile, li2022zero}, so that the optimization can run in real-time.
However, these simplified models usually cannot capture the robot's orientation dynamics accurately, especially when the robot is in the air.
Therefore, it can be difficult to use them for jumping tasks with long periods of air time.
To optimize for more precise jumps, controllers usually need to pre-compute the entire jumping trajectory offline using higher-fidelity models \citep{mitomnidirectional, towr}. 
Compared to these approaches, our method does not require offline optimization,  jumps continuously, and can respond to different landing positions and orientations.

\paragraph{Combining control and learning for legged locomotion} Recently, a number of works have proposed to use reinforcement learning and optimal control jointly for robust and versatile locomotion. 
A common approach is to set up the controller hierarchically, where a high-level policy outputs intermediate commands, which are converted by a low-level controller into motor actions.
While such hierarchical approaches have enabled the robot to walk more efficiently \citep{fast_and_efficient, da2020learning} and conquer difficult terrains \citep{glide}, they are not yet demonstrated on more agile tasks such as jumping.
We use a similar hierarchical setup, but uses a different whole-body controller in the low-level for precise tracking of body acceleration, and achieves continuous, versatile jumping on the real robot.
Another common approach is to use RL to finetune the controller's outputs.
Recently, \cite{bellegarda2020robust} demonstrates that deep RL can improve the quality of a single jump in simulation.
Our work extends their result by using RL to optimize for the controller's performance in the real world, and achieves continuous, omni-directional jumping on the real robot, as well as jump turns.

\section{Overview}
\label{sec:overview}
\begin{figure}[ht]
    \centering
    \includegraphics[width=0.8\linewidth]{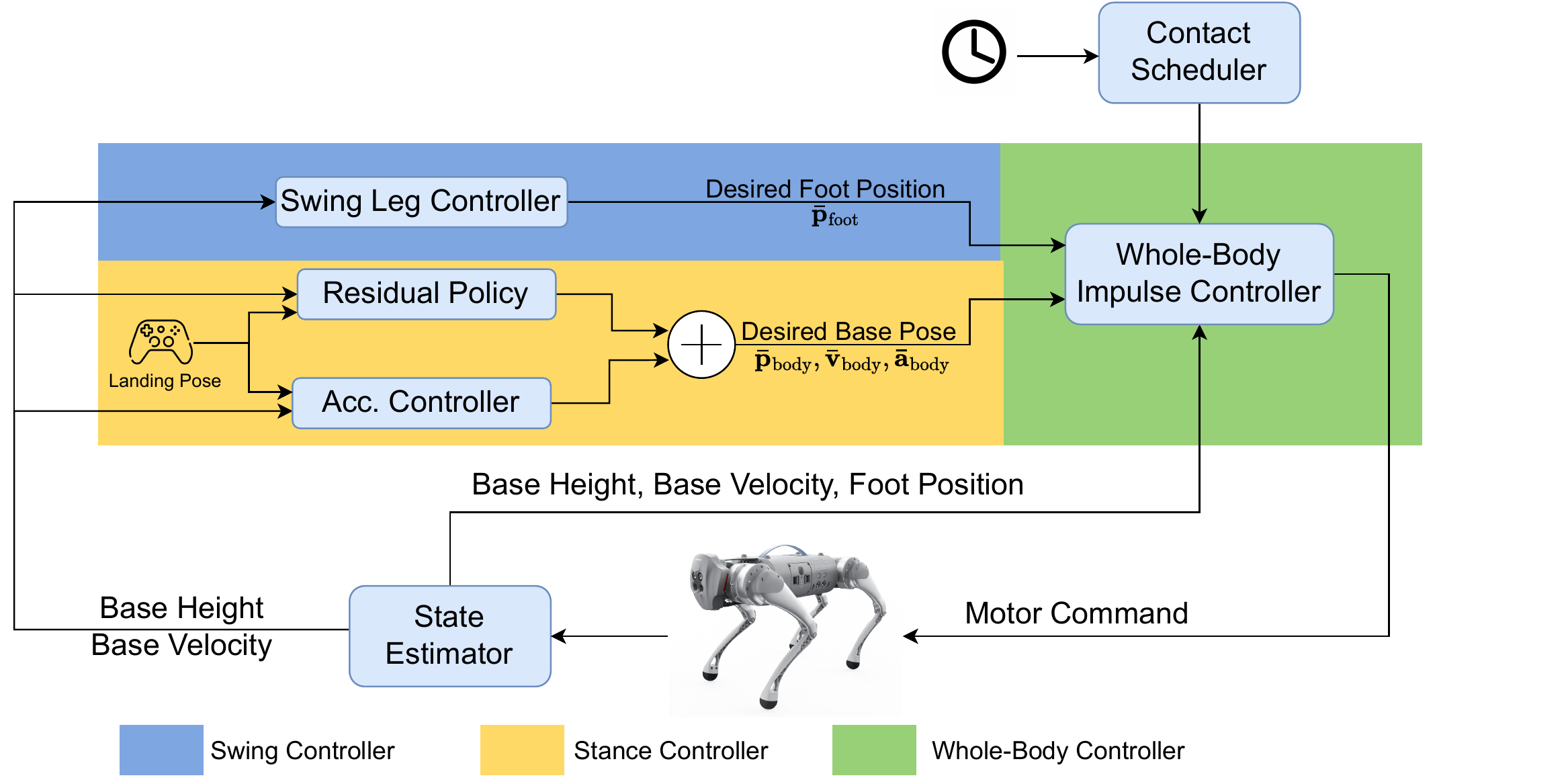}
    \caption{Our hierarchical learning framework controls the jumping of the Go1 robot from two different levels.}
    \label{fig:block_diagram}
\end{figure}

We design a hierarchical framework to learn continuous robot jumping (Fig.~\ref{fig:block_diagram}).
The timing of each jump is regulated by an open-loop time-based \emph{contact scheduler}, which subsequently specifies the desired state of each leg (\emph{swing} or \emph{stance}).
We adopt the pronking gait, where all legs enter and leave the ground at the same time, and set the duration of each jump to be 1 second, which consists of 0.5s of stance time and 0.5s of swing time.
Based on the output from the contact scheduler, we use separate control strategies for swing and stance legs, where the \emph{stance controller} controls the desired body pose and \emph{swing controller} controls the foot positions.
At the low level, a whole-body controller converts the pose command from the swing or stance controller to motor commands.
We also implemented a Kalman Filter based state estimator to estimate the position and velocity of the robot.
We run the entire pipeline at 500Hz so that the robot can respond quickly to external perturbations.

As a critical component of the entire control process, the stance controller needs to achieve sufficient lift-off speed for each jump, while maintaining body stability throughout the entire jump.
To achieve that, we compute the pose command of the stance controller as the sum of a manually designed \emph{acceleration controller} and a learned \emph{residual policy}, where the acceleration controller computes base acceleration to ensure sufficient lift-off velocity based on simplified robot dynamics, and the residual policy fine-tunes the controller's action to ensure stability.
Since the robot is underactuated in the air, we use a simple trajectory controller for swing legs, which computes the desired landing position of each feet according to the Raibert Heuristic \citep{raibertcontroller}.

\section{Learning Residual Policy for Continuous, Versatile Jumping}

\subsection{Reinforcement Learning Preliminaries}
The reinforcement learning (RL) problem is represented as a Markov Decision Process (MDP), which consists of a state space $\mathcal{S}$, an action space $\mathcal{A}$, transition probability $p(s_{t+1}|s_t, a_t)$, reward function: $r:\mathcal{S}\times\mathcal{A}\to\mathbb{R}$, an initial state distribution $p_0(s_0)$ and an episode length $T$. We aim to learn a policy $\pi: \mathcal{S}\to \mathcal{A}$ that maximizes the expected cumulative reward over the entire episode, which is defined as:
\begin{align}
  J(\pi)=\mathbb{E}_{s_0\sim p_0(\cdot), a_t\sim \pi(s_t), s_{t+1}\sim p(\cdot |s_t, a_t)}\sum_{t=0}^T r(s_t, a_t)
\end{align}

\subsection{Environment setup}
We design our environment so that the robot can jump in several directions within each episode. More specifically, each episode consists of 5 jumps, where the robot jumps in-place, 1m forward, 0.5m backward, 0.2m to the left, and 0.2m to the right, in that order. 
Before each jump, the environment records the robot's current position and computes the desired landing position based on the jumping directions. 
This information is then supplied to the residual policy to compute the desired pose commands, which is subsequently sent to the low-level whole-body controller. The details of our environment are as follows:

\paragraph{State Space} The state space includes the robot state and task information. More specifically, the robot state includes the position, orientation, linear velocity, and angular velocity of the base, as well as the foot positions. The task information includes the robot's distance to the desired landing position, and the remaining time in the current locomotion cycle.

\paragraph{Action Space} \label{section:action_space}
Since the low-level whole-body controller is effective for precise tracking of the body pose \citep{mit_wbic}, we directly command the desired body pose in our environment.
In the original work by \cite{mit_wbic}, the whole-body controller takes in an 18-dimensional vector specifying the position, velocity, and acceleration for each of the base's 6 DoFs.
To reduce the search dimension, we design the action space to specify one command for each DoF of the base, and computes the other two dimensions of each DoF heuristically.
More specifically, the action space specifies the desired linear \emph{acceleration} (3-dimensional) and angular \emph{acceleration} around the $z$ axis (1-dimensional), so that the policy can directly control the planar and vertical movements of the body, as well as turning, which can change rapidly within each jump.
The action space specifies the desired \emph{position} for the remaining two DoFs, namely the body roll and pitch, to avoid unnecessary body oscillations.
The remaining commands for each DoF is specified heuristically. See section.~\ref{sec:wbc} for more details.

\paragraph{Reward} We design the reward function as the linear combination of alive bonus, distance penalty, orientation penalty and contact consistency penalty:

\begin{align}
	r=r_{\text{alive}}+w_p\cdot r_{\text{position}} + w_o \cdot r_{\text{orientation}} + w_c\cdot  r_{\text{contact}}
\end{align}
where the alive bonus, $r_{\text{alive}}=4$, is a fixed constant that makes the total return positive.
The position reward, $r_{\text{position}}$ is the squared distance between the robot's current position and the desired landing position, normalized by the total desired jumping distance. The orientation reward, $r_{\text{orientation}}=-(\text{roll}^2+\text{pitch}^2)$ encourages the robot to stay upright. The contact consistency reward $r_{\text{contact}}=\sum_{i=1}^4  \mathbbm{1}(c_i\neq \hat{c}_i)$ is the sum of 4 indicator functions about the contact situation of each leg, where $\hat{c}_i\in \{0, 1\}$ is the desired contact of foot $i$ according to contact scheduler (Section.~\ref{sec:overview}) and $c_i\in\{0, 1\}$ is the actual contact of foot $i$. Intuitively, the contact consistency reward ensures that the robot jumps according to the desired schedule without significant mismatch in lift-off and touch-downs.
We use $w_p=1, w_o=5, w_c=0.4$ in all our experiments.

\paragraph{Early Termination} To prevent the learning algorithm from unnecessary explorations in suboptimal regions, we terminate an episode early if the robot falls (i.e. when the base height is lower than 8cm, when the cosine distance between body's upright vector and the gravity direction is less than 0.6, or when any body parts came in contact with the ground).

\subsection{Manually-designed Acceleration Controller}
Due to the discrete contact change, the reward landscape in the jumping environment can be highly non-smooth with local minima, which makes it challenging to learn using standard exploration strategies in RL algorithms. 
To facilitate learning, we manually design an acceleration controller as the base policy for the environment, and uses reinforcement learning to learn \emph{residual actions} to finetune the policy's performance. The acceleration controller models the robot body as a single point mass, computes the desired lift-off velocity based on contact timing, and tracks this lift-off velocity using simple heuristics.

\paragraph{Computing the Lift-off Velocity} The acceleration controller computes the desired lift-off velocity $\vec{v}_{\text{liftoff}}=(v_x, v_y, v_z, v_{\text{yaw}})$ based on the desired landing displacement $p_x, p_y, p_{\text{yaw}}$ and the swing time $t_{\text{swing}}$. The planar velocities, $v_x=\frac{p_x}{t_{\text{swing}}}, v_y=\frac{p_y}{t_{\text{swing}}}, v_{\text{yaw}}=\frac{p_{\text{yaw}}}{t_{\text{swing}}}$, is the average flying speed required for the robot to land at the desired position and orientation. The vertical velocity, $v_z=\frac{1}{2} g t_{\text{swing}}$, is the minimum vertical velocity for the robot to maintain the desired swing time, where $g$ is the gravity constant.

\begin{figure}[t]
    \centering
    \includegraphics[width=1\linewidth]{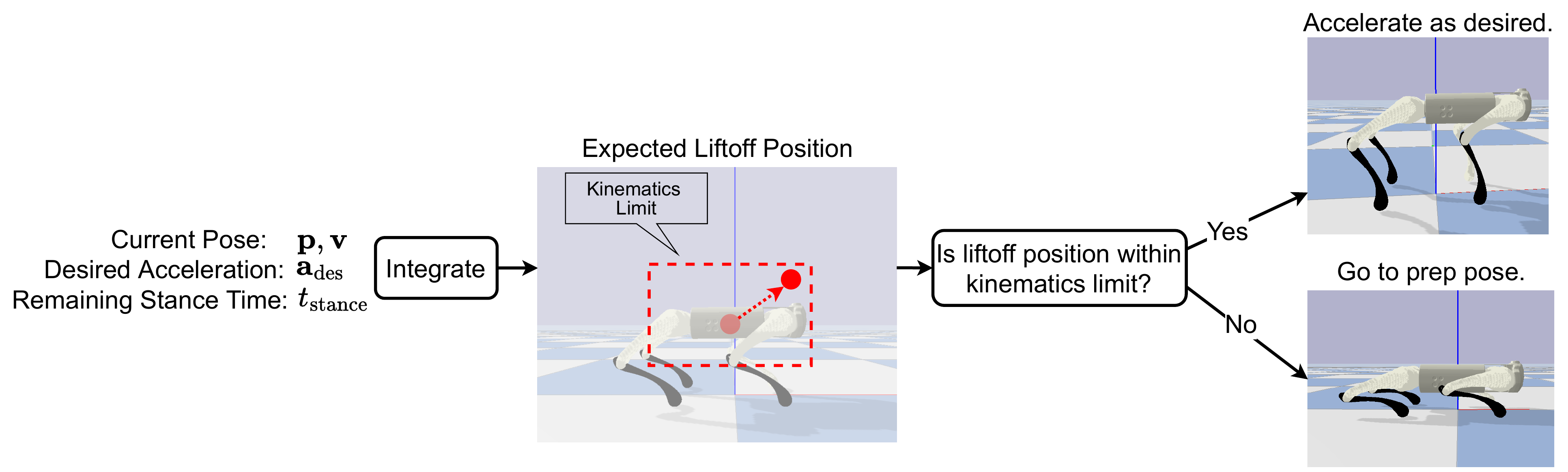}
    \caption{The acceleration controller decides whether to execute the desired acceleration command $\vec{a}_{\text{des}}$ based on the estimated lift-off position, which is computed by numerical integration. If the lift-off position (dark red dot) is within the approximated kinematics limit (dashed square), the controller executes the $\vec{a}_{\text{des}}$. Otherwise the controller moves the robot to a low-standing preparation pose.}
    \label{fig:jump_controller}
\end{figure}

\paragraph{Tracking the Lift-off Velocity} To track the desired lift-off velocity $\vec{v}_{\text{liftoff}}$, the acceleration controller computes the desired acceleration $\vec{a}_{\text{des}}=\frac{\vec{v}_{\text{liftoff}}-\vec{v}}{t}$ based on the remaining time in the stance phase $t$, the desired liftoff velocity $\vec{v}_{\text{liftoff}}$ and the current CoM velocity $\vec{v}$.
The acceleration controller then either executes this acceleration $\vec{a}_{\text{des}}$, or moves the robot to a preparation position, based on an estimation of the lift-off position (Fig.~\ref{fig:jump_controller}).
More specifically, the acceleration controller assumes the robot as a point-mass, and computes the body's CoM position at lift-off time based on the current pose and desired accelerations.
If the lift-off position violates kinematics limits (e.g. if the base is so high that foot contact cannot be maintained), the controller computes an alternative acceleration that moves the robot to a low-standing preparation position.
Otherwise, the controller outputs the desired acceleration $\vec{a}_{\text{des}}$.
To simplify computation, we approximate the feasible CoM positions as a bounding box around the robot's current CoM.

\subsection{Training the Residual Policy}
To improve the performance of the acceleration controller, we train a policy to add an action residual to the acceleration controller's outputs.
We represent our policy as a neural network with 1 hidden layer of 256 units and tanh activation.
We train our policy using Augmented Random Search (ARS) \citep{ars}, a simple, parallelizable evolutionary algorithm that has been successfully applied to locomotion tasks \citep{pmtg, tan2011articulated, tan2016simulation, geijtenbeek2013flexible}.
We chose ARS because it explores in the policy parameter space and does not directly inject noise in action outputs.
Moreover, ARS evaluates policies require accurate estimation of the value function, which can be challenging for hierarchical tasks \citep{fast_and_efficient} due to its non-Markovian nature.

\section{Low-level Whole-body Controller}
At the low-level, we use a whole-body controller (WBC) to convert the body and foot pose commands into motor actions. Our implementation is based on the work by \cite{mit_wbic} with a few modifications. We briefly summarize the controller design here. 
Please refer to the original work for further details.

\paragraph{Interface with Stance Controller}
\label{sec:wbc}In summary, WBC takes in a 18-dimensional vector that specifies the desired pose $\vec{p}_{\text{body}}$, velocity $\vec{v}_{\text{body}}$, and acceleration $\vec{a}_{\text{body}}$ for each of the 6 DoFs of the robot \emph{body}, as well as the foot swing positions $\vec{p}_{\text{foot}}$.
The foot swing positions $\vec{p}_{\text{foot}}$ is directly specified by the swing controller (Section.~\ref{sec:overview}).
The stance controller (Section.~\ref{section:action_space}) specifies 4 out of the 6 dimensions for the base accelerations $\vec{a}_{\text{body}}$ (3 linear accelerations and angular accelerations around the $z$ axis), as well as 2 out of the 6 dimensions of the base pose $\vec{p}_{\text{body}}$ (roll and pitch). For the remaining pose commands, we set the desired body pose $p_{\text{body}}$ to be the current body pose, the desired linear body velocity to the current velocity, the desired angular body velocity to 0, and the desired angular acceleration around the $x, y$ axis to 0.

\paragraph{Computation of Motor Commands}WBC computes an impedance command that specifies the desired position $\bar{\vec{q}}$, velocity $\bar{\dot{\vec{q}}}$ and torque $\bar{\vec{\tau}}$ for each \emph{motor}.
The applied torque $\vec{\tau}$ is the sum of the desired torque plus the PD feedback:
\begin{equation}
    \vec{\tau} = k_p (\bar{\vec{q}}-\vec{q})+k_d(\bar{\dot{\vec{q}}}-\dot{\vec{q}})+\bar{\vec{\tau}}
\end{equation}
where $\vec{q}, \dot{\vec{q}}$ is the current position and velocity of each motor, and $k_p, k_d$ are fixed gains. To compute the motor command, WBC first applies an inverse kinematics algorithm, which computes the desired position $\bar{\vec{q}}$ and velocity $\bar{\dot{\vec{q}}}$ for each motor, in order to move the robot to the desired body position $\vec{p}_{\text{body}}$ and velocity $\vec{v}_{\text{body}}$. After that, WBC computes the additional motor torque $\bar{\vec{\tau}}$ required to achieve the desired base accelerations $\vec{a}_{\text{body}}$, based on the full rigid-body dynamics model of the robot.

\section{Results}
\subsection{Experiment Setup}
We test our framework on a Go1 robot from Unitree \citep{go1robot}, which is a small-scale, 15kg quadrupedal robot with 12 degrees of freedom.
We build the simulation environment of Go1 using Pybullet \citep{pybullet}, and implement the entire control pipeline, including the the state estimator, low-level WBC, the jump controller and the residual policy in Python.
We train the residual policy on a desktop computer with a 16-core CPU, where the training takes around 3 hours to complete.
\subsection{Continuous, Versatile Jumping}
\begin{figure}[t]
    \centering
    \begin{subfigure}[t]{0.48\linewidth}
        \includegraphics[width=\linewidth]{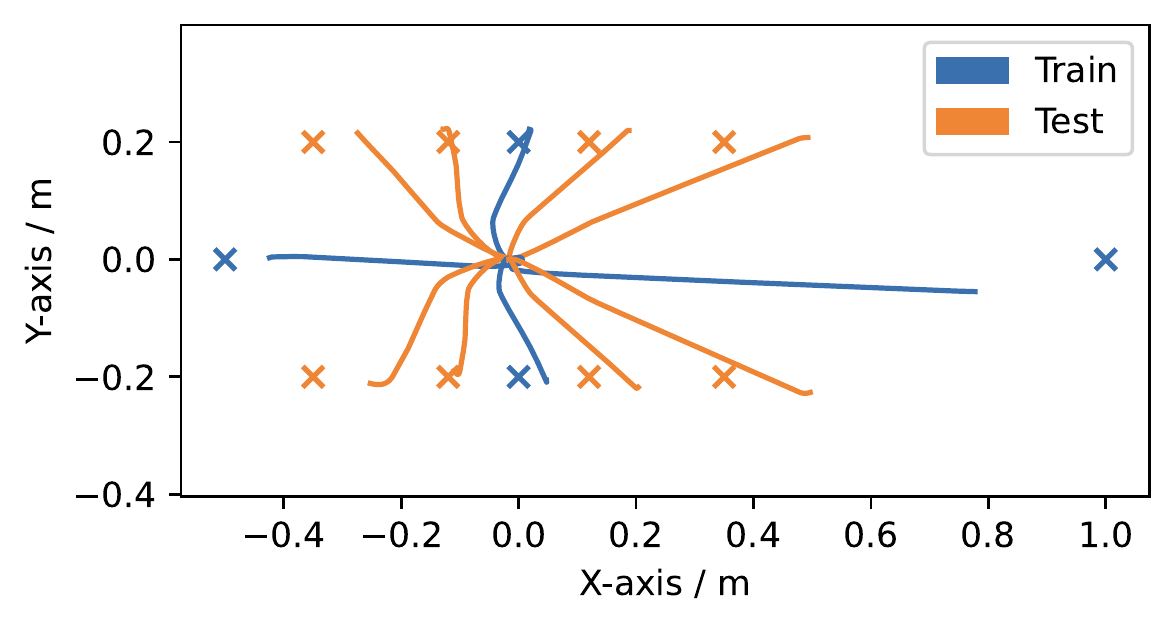}
        \caption{Omni-directional jumping: CoM trajectory}
        \label{fig:omni_jump}
    \end{subfigure}
    \begin{subfigure}[t]{0.45\linewidth}
        \includegraphics[width=\linewidth]{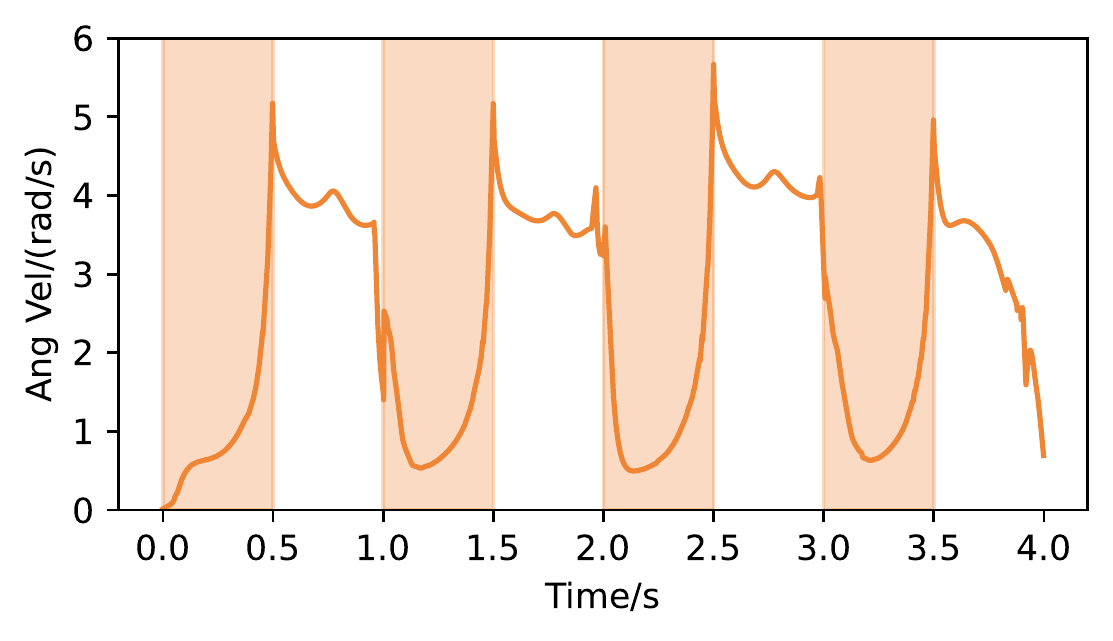}
        \caption{Jump-turn: $z$-component of angular velocity }
        \label{fig:jump_turn}
    \end{subfigure}
    \caption{Different jumping skills learned by our framework. \textbf{Left}: Omni-directional jumping. Lines show birds-eye view of CoM trajectory, where the robot starts facing positive $x$ direction. Crosses show desired landing positions. Colors show whether the direction is seen during training. \textbf{Right}: Continuous jump-turn. Plot shows the z-component of angular velocity (approximately the change rate of yaw angle). Shaded area indicates foot contact.}
    \label{fig:generalization}
\end{figure}

We test the performance of our learned framework on a series of jumping tasks, where each task specifies either a different desired jumping direction, or a desired turning rate (Fig.~\ref{fig:generalization}).
Although we train the framework in only 4 jumping directions, the resulting controller interpolates between them and jumps in intermediate directions (fig.~\ref{fig:omni_jump}).
The framework also enables the robot to jump and turn \emph{simultaneously}, and achieves an average turning rate of about 3.5 rad/s (fig.~\ref{fig:jump_turn}). 
For omni-directional jumping, we also notice that the controller tends to overshoot for forward jumps, and undershoot for backward jumps.
This is likely due to the asymmetric leg designs of the robot platform, which generates higher accelerations in the forward direction than in the backward direction.

\subsection{Transfer to the real robot}
\begin{figure}[th]
    \centering
    \includegraphics[width=0.85\linewidth]{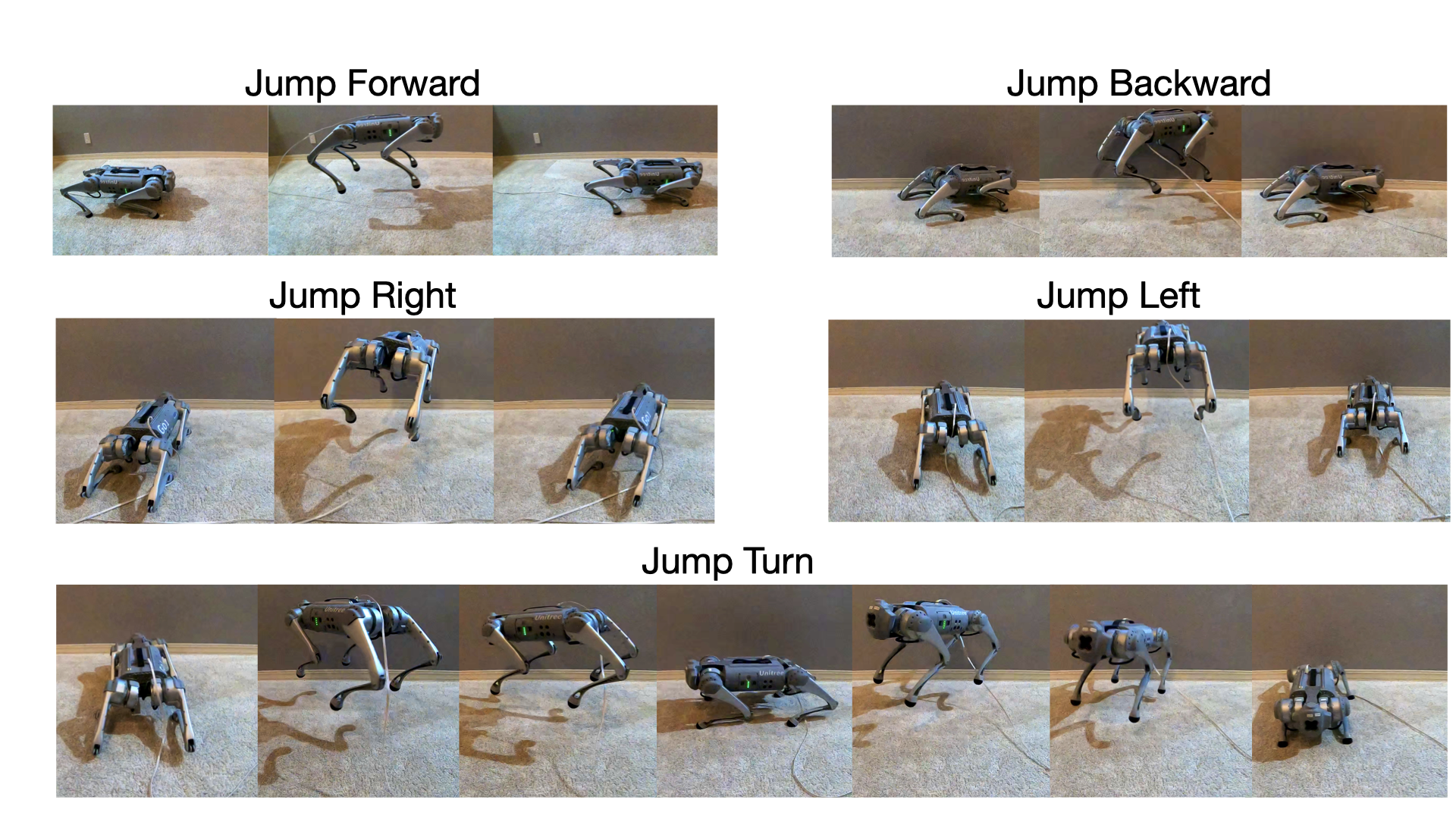}
    \caption{Footages of the Go1 robot completing several jumps in the real world.}
    \vspace{-1em}
    \label{fig:real_robot}
\end{figure}

We deploy the learned residual policy directly to the real world without additional finetuning. Thanks to the robustness of the low-level WBC, our framework can complete several high jumps in the real world, including jumping in multiple directions and turning (Fig.~\ref{fig:real_robot}). Please visit the website for videos. The robot achieves a maximum jumping height of around 50cm, a forward jumping distance of around 60cm, and a maximum turning rate of 90 degrees per jump. Note that this jumping performance in the real world is slightly lower than the robot's performance in simulation. We hypothesize this as a result of unmodeled motor saturation, and plan to investigate further in future works.

\subsection{Comparison with baseline policies}
\begin{figure}
    \centering
    \includegraphics[width=0.7\linewidth]{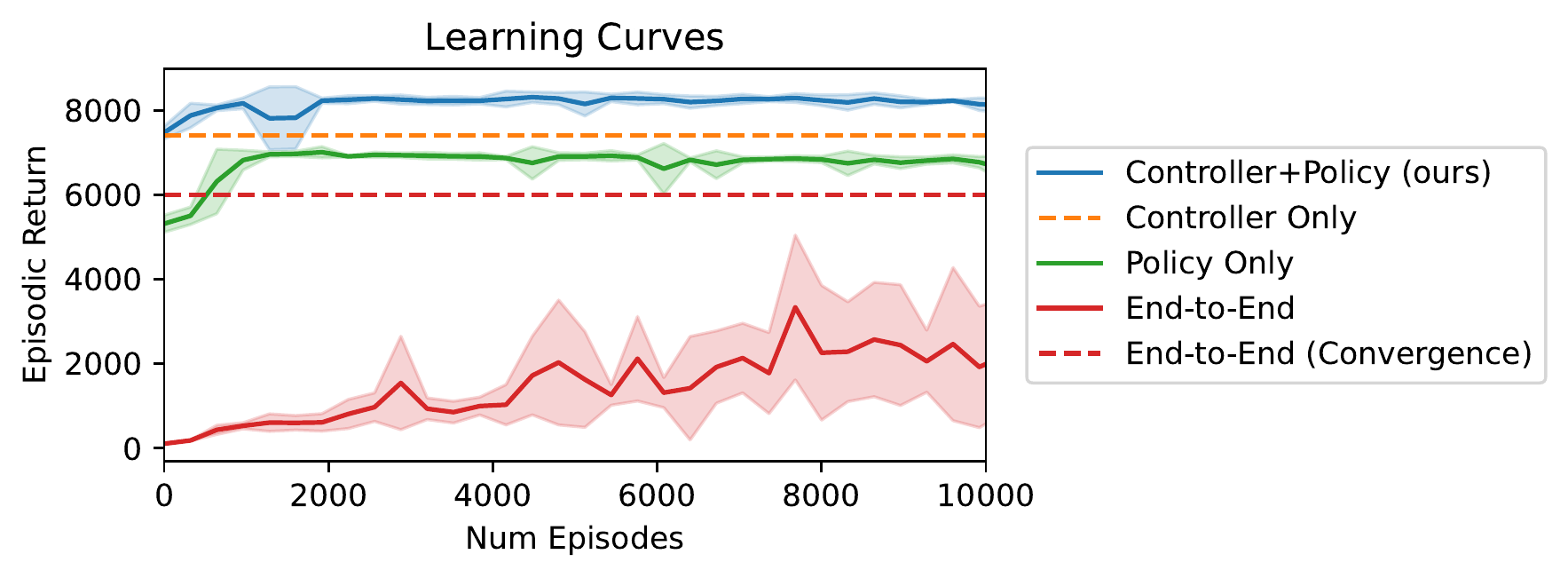}
    \vspace{-1em}
    \caption{Learning curves of our framework compared with baseline methods. Error bar shows 1 standard deviation.}
    \vspace{-1em}
    
    \label{fig:learning_curves}
\end{figure}

To further validate our design choices, we conduct an ablation study by removing either the residual policy or the acceleration controller from our pipeline. We also compare our framework with an end-to-end RL policy that directly outputs motor position commands. The result is summarized in figure.~\ref{fig:learning_curves}.

\begin{figure}[t]%
  \begin{minipage}[b]{0.48\linewidth}
    \small
        \centering
    \begin{tabular}{c|c|c}
    \hline
    \textbf{Task} & \textbf{Controller} & \textbf{Controller}\\
    & \textbf{+ Policy} & \textbf{Only}\\\hline
    Forward &100\%&20\%\\
    Backward &80\%&0\%\\
    Left &100\%&80\%\\
    Right &100\%&60\%\\
    Jump-Turn &100\%&0\%\\\hline
    \end{tabular}
    \vspace{1.3em}
    \captionof{table}{Success rates (over 5 trials) with or without the residual policy. A jump is successful if it does not trigger the early termination condition (Sec.~\ref{section:action_space}).}
    \label{table:success_rates}
  \end{minipage}
  \hfill
  \begin{minipage}[b]{0.50\linewidth}
    \centering
    \includegraphics[width=\linewidth]{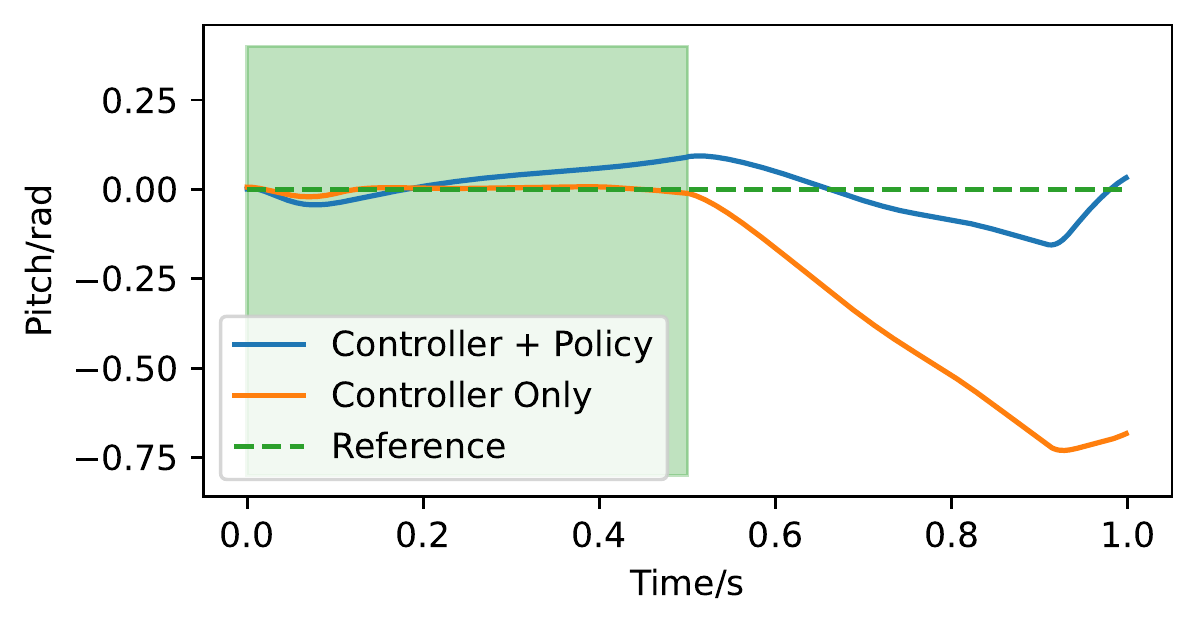}
    \vspace{-2em}
    \captionof{figure}{Robot pitch angle during a backward jump using different controllers. Shaded area shows stance phase.}
    \label{fig:pitch_comparison}
  \end{minipage}
\end{figure}

\paragraph{Acceleration Controller Only} We test the performance of the manually designed acceleration controller without learned residual policy on the same jumping task. While the controller completes all the jumps without falling over, it achieves a lower reward without the residual policy (Fig.~\ref{fig:learning_curves}). We also find that the residual policy increases the overall success rate of the robot in the real world (Table.~\ref{table:success_rates}).
To further understand the effectiveness of the residual policy, we perform a backward jump with and without the residual policy, and plot the base pitch angle in Fig.~\ref{fig:pitch_comparison}. 
While the acceleration controller maintains the body pitch closer to reference in the stance phase, it causes significantly larger pitch angle deviations in the swing phase. 
This is because the acceleration controller approximates the robot as a point-mass, and cannot account for changes in body orientations. 
In contrast, the residual policy learns to correct this pitch angle deviation by slightly shifting the body pitch in stance phase, which results in lower overall pitch deviations throughout the entire jump. 

\paragraph{Policy Only}
\begin{figure}[t]
    \centering
    \begin{subfigure}[t]{0.4\linewidth}
        \includegraphics[width=\linewidth]{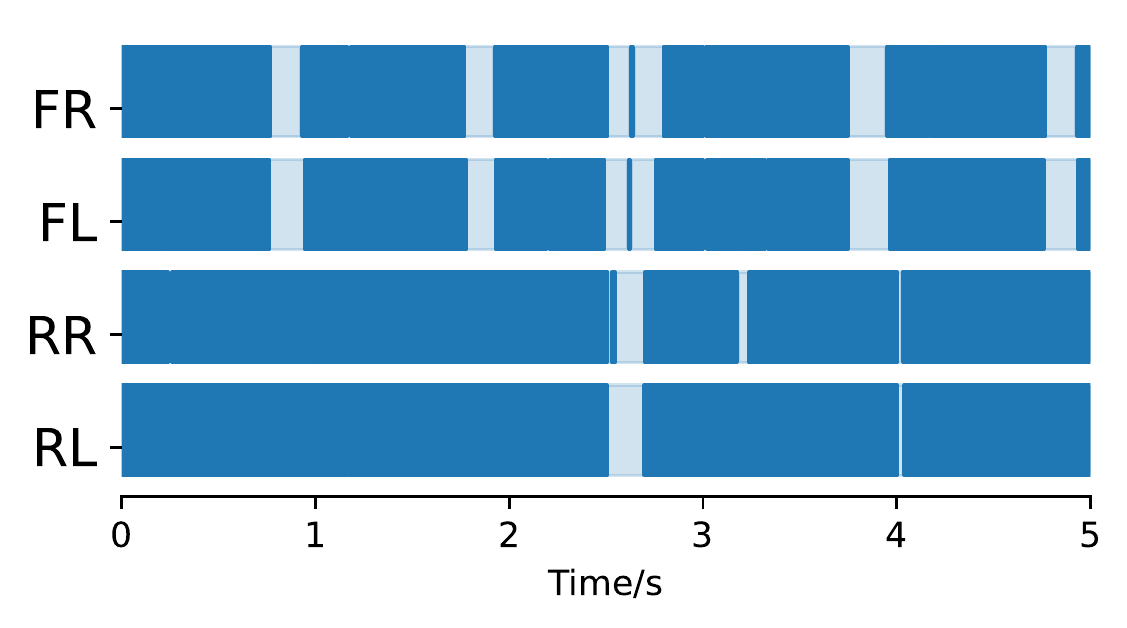}
        \vspace{-2em}
        \caption{Policy Only.}
        \label{fig:contact_nores}
    \end{subfigure}
    \begin{subfigure}[t]{0.4\linewidth}
        \includegraphics[width=\linewidth]{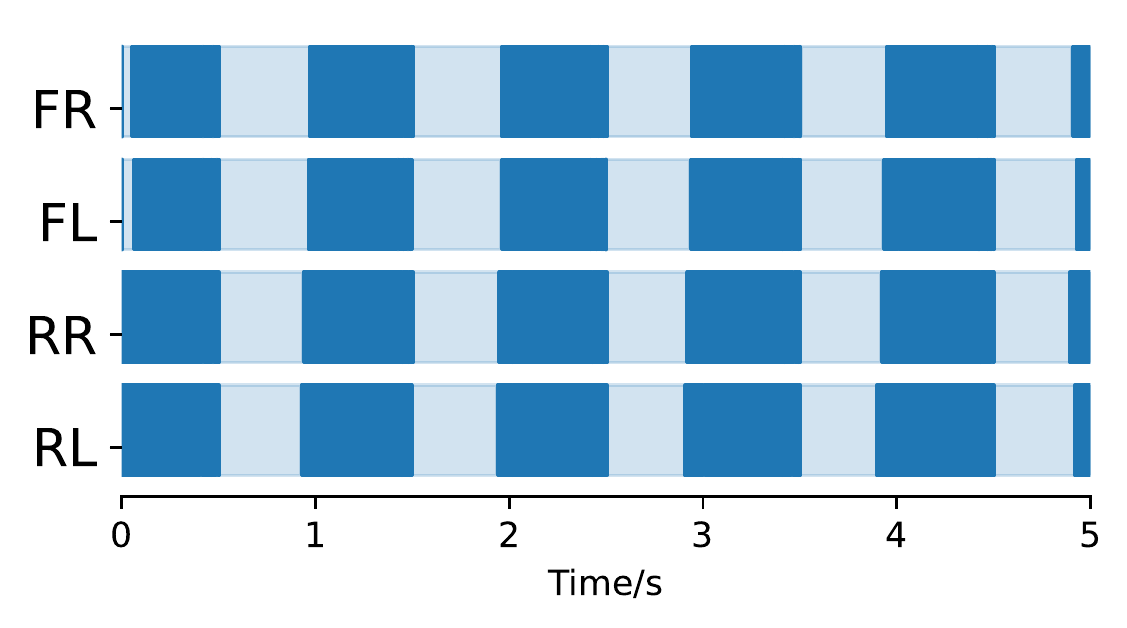}
        \vspace{-2em}
        \caption{Controller + Policy (ours).}
        \label{fig:contact_res}
    \end{subfigure}
    \vspace{-0.5em}
    \caption{Foot contacts for controllers trained without and with the jump controller. Dark area indicates foot contact.}
    \vspace{-1em}
    \label{fig:generalization}
\end{figure}
Without the acceleration controller, the residual policy gets stuck in a local minima, and achieves a low reward (Fig.~\ref{fig:learning_curves}).
The resulting policy does not achieve sufficient height for each jump, and keeps legs in contact most of the times (Fig.~\ref{fig:contact_nores}).
We hypothesize this as a result of the frequent contact changes in the environment, which can create a noisy reward landscape for the algorithm to optimize.
In contrast, the policy trained with the controller achieves consistent, high flight times for each jump (Fig.~\ref{fig:contact_res}).

\paragraph{End-to-End RL} Lastly, we compare our method to a baseline, where we train an end-to-end reinforcement learning policy that directly outputs motor position commands. Similar to previous works \citep{rma, eth_adaptation,sim-to-real}, we reduce the control frequency to 50Hz to avoid high-frequency motor oscillations. We use the same state space and reward function in the environment design, and train the policy using the same ARS algorithm.
We find that the E2E policy learns significantly slower, and requires around 10 times more training episodes (Fig.~\ref{fig:learning_curves}). 
The policy also achieves the lowest overall reward compared to the other policies.
While additional efforts in reward shaping and imitation learning \citep{laikago_imitation} can improve the performance of the E2E policy, our method creates a simple, data-efficient alternative that does not require pre-recorded demonstration trajectories.
\section{Conclusion}
In this work, we present a hierarchical framework to for continuous quadruped jumping, which consists of a manually designed acceleration controller, a learned residual policy, and a low-level whole-body controller.
The trained framework can be transferred directly to the real world, and is capable of continuous jumping at arbitrary angle and distances according to user specification, as well as jump-turning.
One limitation of our work is that it currently only supports the pronking gait, where all 4 legs leave or touch the ground at the same time.
Supporting more versatile gaits such as bounding or galloping can potentially increase the height and distance of the jump, which we plan to investigate in future work.
Another future direction is to integrate perception into our framework, so that the robot can use its jumping skills to traverse through difficult terrains autonomously.

\clearpage
\bibliography{reference}

\end{document}